\documentclass[runningheads]{llncs}

\usepackage[year=2024]{eccv}
\usepackage{eccvabbrv}

\usepackage{graphicx}
\usepackage{booktabs}

\usepackage[accsupp]{axessibility}  

\usepackage[pagebackref,breaklinks,colorlinks,citecolor=eccvblue]{hyperref}

\usepackage{orcidlink}

\begin{document}

\title{Scaling Multi-Camera 3D Object Detection through Weak-to-Strong Eliciting} 

\titlerunning{Scale BEV}

\author{Hao LU$^{1,2}$, Jiaqi TANG$^{1,2}$, Xinli XU$^{1,2}$, Xu CAO$^{1,2}$, Yunpeng ZHANG$^3$, Guoqing WANG$^4$, Dalong DU$^3$, Hao CHEN$^{2}$, Yingcong CHEN$^{1,2}$ \\}

\authorrunning{Scale BEV}
\institute{$^{1}$The Hong Kong University of Science \& Technology (Guangzhou), \\
$^{2}$The Hong Kong University of Science \& Technology, $^{3}$ PhiGent Robotics,\\
$^{4}$ Shanghai Jiao Tong University
}

\maketitle

\begin{abstract}

The emergence of Multi-Camera 3D Object Detection (MC3D-Det), facilitated by bird's-eye view (BEV) representation, signifies a notable progression in 3D object detection. Scaling MC3D-Det training effectively accommodates varied camera parameters and urban landscapes, paving the way for the MC3D-Det foundation model. However, the multi-view fusion stage of the MC3D-Det method relies on the ill-posed monocular perception during training rather than surround refinement ability, leading to what we term "surround refinement degradation". To this end, our study presents a weak-to-strong eliciting framework aimed at enhancing surround refinement while maintaining robust monocular perception. Specifically, our framework employs weakly tuned experts trained on distinct subsets, and each is inherently biased toward specific camera configurations and scenarios. These biased experts can learn the perception of monocular degeneration, which can help the multi-view fusion stage to enhance surround refinement abilities. Moreover, a composite distillation strategy is proposed to integrate the universal knowledge of 2D foundation models and task-specific information. Finally, for MC3D-Det joint training, the elaborate dataset merge strategy is designed to solve the problem of inconsistent camera numbers and camera parameters. We set up a multiple dataset joint training benchmark for MC3D-Det and adequately evaluated existing methods. Further, we demonstrate the proposed framework brings a generalized and significant boost over multiple baselines. Our code is at \url{https://github.com/EnVision-Research/Scale-BEV}.  

\end{abstract}

\section{Introduction}

Multi-Camera 3D Object Detection (MC3D-Det) is a critical technology for detecting and localizing objects within a surrounding space through an array of cameras, effectively synthesizing data from multiple viewpoints to achieve more accurate and robust detection~\cite{ma2022vision,li2022bevsurvey}. This is particularly beneficial in environments where objects may be occluded or poorly visible from certain angles. MC3D-Det operates in two primary phases: initially, image features are extracted from individual camera perspectives (monocular perception), followed by the amalgamation of these features to formulate a final prediction (multi-view fusion). The Bird's-Eye View (BEV) approach, in particular, is renowned for its clarity and effectiveness in navigating the complexities of MC3D-Det tasks~\cite{huang2021bevdet,li2022bevformer,lin2022sparse4d,lin2023sparse4dv3,li2022bevdepth,FBBEV}.

Current MC3D-Det models, often trained within a traditional single-dataset framework, tend to overfit specific camera configurations and environmental variables. This limitation becomes glaringly apparent when deploying these models across diverse datasets, where they typically perform at merely about 50$\%$ of their potential compared to their performance within the original domain, across various camera setups~\cite{Wang_2023_CVPR,lu2023pdbev}. Such a discrepancy underscores the algorithms' restricted versatility across different vehicle types, which might differ significantly in terms of camera placement and model.

\begin{figure}[t]
    \centering
    \includegraphics[width=1\linewidth]{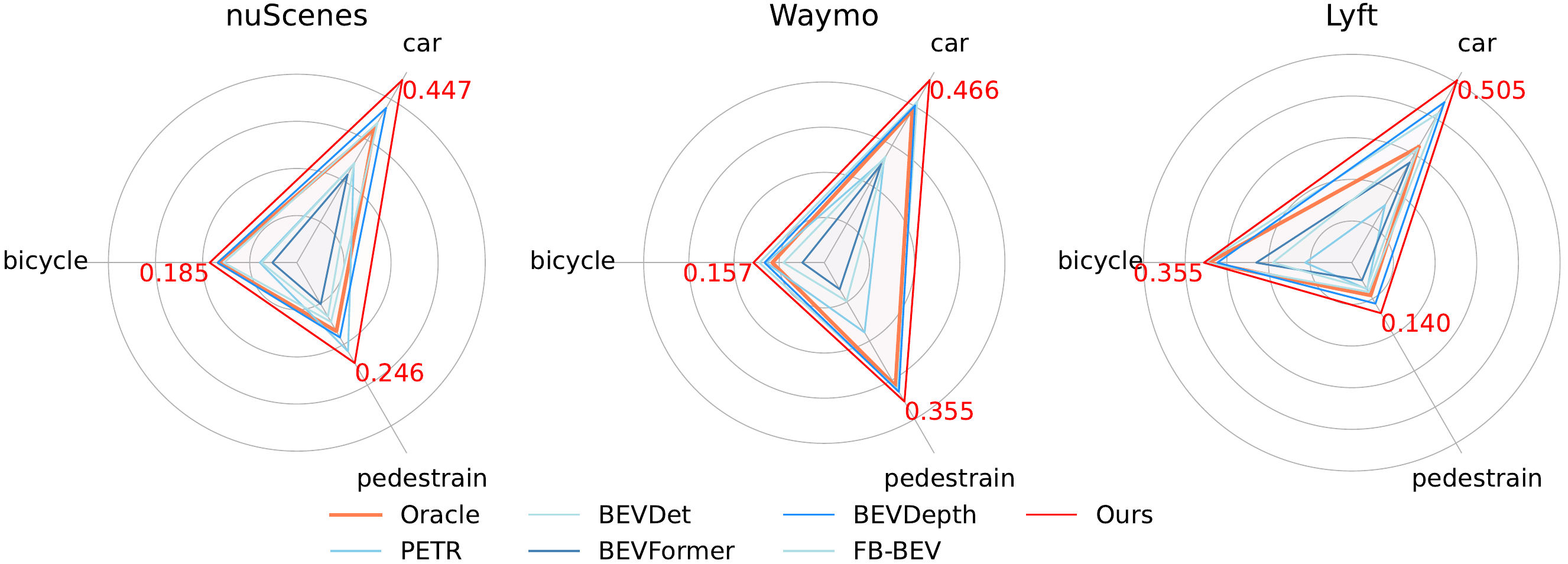}
    \caption{Results of different methods on combined datasets. Oracle denotes the results obtained from the evaluations where models are trained and validated on the same dataset. }
    \vspace{-4mm}
    \label{fig:leida}
\end{figure}

Addressing this challenge, employing a multi-dataset training strategy for MC3D-Det algorithms has shown promise. This method starts with combining various data sets and retraining the baseline detector, including DETR3D~\cite{wang2022detr3d}, PETR~\cite{liu2022petr}, BEVFormer~\cite{li2022bevformer}, BEVDet~\cite{huang2021bevdet}, BEVDepth~\cite{li2022bevdepth}, FB-BEV~\cite{FBBEV} as illustrated in Fig.~\ref{fig:leida}. It is imperative to highlight that existing algorithms do not achieve significant advantages over Oracle. Increasing the number and diversity of data sets does not lead to greater performance gains.

Our further exploration unleashes the \emph{surround refinement degeneration} phenomenon, whereby the reliance on monocular depth prevents the multi-view fusion model from learning surround refinement capacity. Specifically, the monocular depth estimation effective within training datasets, significantly deteriorates during testing phases without understanding different camera parameters and scenes. To make matters worse, during training, the multi-view fusion model tends to rely on the accurate monocular perception process instead of refining geometric errors by fusing different views, called surround refinement degradation.

In this paper, we design a weak-to-strong eliciting MC3D-Det framework that significantly improves both monocular perception and surround refinement capabilities. Specifically, we design a composite distillation strategy to learn large-scale image knowledge from the 2D pre-trained foundation model, which also establishes auxiliary tasks to inject depth information and more fine-grained information. 
This effectively improves the monocular perception ability, which is critical for the overall accuracy of MC3D-Det. Yet this also introduces a side effect as mentioned above, i.e., potential overreliance on monocular cues and overlook of the surround refinement ability of the multi-view fusion module. As a result, the model may not generalize well on new camera parameters due to the degradation of monocular perception. To improve the ability to surround refinement while preserving monocular perception, we trained multiple weakly tuned experts on distinct sub-datasets. During training, these experts simulate ill-posed monocular features encountered in different camera settings and scenes. These ill-posed monocular features assist the multi-view fusion model in using multi-view information to refine geometric information, thereby augmenting the surround refinement ability. To better complete the multi-dataset joint training of the MC3D-Det task, we designed a delicate merge strategy to solve the problem of inconsistent camera parameters and camera numbers. Finally, we built a benchmark for multi-datasets joint training and sim-real joint training. In summary, our paper contributions are:

\begin{itemize}

    \item A plug-and-play MC3D-Det framework is designed to accommodate various camera parameters and scenes, significantly improving both monocular perception and surround refinement capabilities.
    
    \item The composite distillation is designed not only to leverage more generalized semantic representations of 2D foundation models while providing additional fine-grained learning supervision.
   
    \item We build the first benchmark for joint training of multiple datasets and create a general joint training recipe for the MC3D-Det task, demonstrating our method's efficacy. 
    
    \item By applying our framework to various MC3D-Det algorithms, we showcase its wide applicability and efficiency, without any additional inference time costs.
    
\end{itemize}

\section{Related Works}

\subsection{Vision-based 3D object detection}

The field of Multi-camera 3D object detection (MC3D-Det), which aims to identify and localize objects in 3D space, has garnered significant attention as evidenced by recent studies~\cite{ma2022vision,li2022bevsurvey}. Current MC3D-Det methods primarily involve extracting image features and projecting them onto a bird's-eye view (BEV) plane, enhancing the integration of spatial-temporal features. Initial forays into this area, exemplified by Orthographic Feature Transform (OFT) and Lift-splat-shoot (LSS), have proven the viability of transposing multi-view features into BEV space~\cite{roddick2018orthographic,philion2020lift,huang2021bevdet,li2022bevdepth}. Recent works have explored the application of transformer technology to understand 2D and 3D mapping relationships~\cite{li2022bevformer,yang2023bevformerv2}. Additionally, FB-BEV has endeavored to integrate transformer and LSS-based methods, employing both forward and backward mechanisms~\cite{FBBEV}.  The PETR series, introducing 3D position-aware encoding, enables networks to implicitly assimilate geometric information~\cite{liu2022petr,liu2022petrv2}. The SparseBEV family, utilizing a sparsity query, aims to enhance both the speed and accuracy of models ~\cite{liu2023sparsebev,lin2022sparse4d,lin2023sparse4d,lin2023sparse4dv3}. Although these approaches yield satisfactory outcomes on in-distribution datasets, their effectiveness is considerably diminished under cross-domain conditions, even with specialized designs~\cite{Wang_2023_CVPR,lu2023pdbev}. Consequently, integrating diverse datasets is essential to adapt to varying camera parameters and scenarios for the MC3D-Det task.

\subsection{Multiple dataset joint training}

Multi-dataset joint training has been explored in the image domain such as classification, detection, and segmentation~\cite{kirillov2023sam,yin2023metric3d,liang2022multi,yuan2023ad,hao2023language,zhou2023uni3d,wu2023towards}. 
In most cases, combining multiple data sets can greatly improve the final performance of the model~\cite{kirillov2023sam,yin2023metric3d,liang2022multi,yuan2023ad,wu2023towards}, which is due to learning more general and robust features. There has also been some work that has found that simply merging data sets does not improve performance. For example, Uni3D found that the point cloud sparsity of different data sets and the attribute distribution of objects were different, so a series of schemes were adopted to improve the performance of 3D point cloud detection~\cite{zhang2023uni3d}. Some work attempts to address the inherent differences and labeling of different data sets in point cloud understanding~\cite{wu2023towards,liang2023label}. TMT-VIS uses taxonomic guidance to utilize multiple video instance segmentation datasets effectively~\cite{zheng2023tmt}. Unlike these tasks, MC3D-Det requires enhanced robustness of the model to internal and external camera parameters on different datasets.

\section{Problem Definition}

\subsection{Problem Paradigm}

This study aims to improve the effectiveness of MC3D-Det by integrating multiple datasets and leveraging the pre-training of 2D base models. We consider the $i$-th dataset $D_i=\{X^i, Y^i, K^i, E^i\}$, comprising numerous samples. The $j$-th sample of the $i$-th dataset encompasses $N$ multi-view images $X^i_j=\{I_1, I_2, ..., I_N\}^i_j$, associated labels $Y^i_j$, and the intrinsic $K^i_j$ and extrinsic parameters $E^i_j$ of the multi-view cameras. 

Joint training with multiple datasets is designed to enhance MC3D-Det model adaptability across varied scenes and camera configurations. The MC3D-Det approach entails three foundation processes: (1) the monocular perception encoder $F_{mono}$ extracts features from multiple datasets and pre-trained 2D models $G$; (2) the surround refinement model $F_{fusion}$ (or name multi-view fusion model) integrates and refines these image features. (3) the detector $D$ predicts the position and properties of objects by the surrounding features. Our objective is to train the monocular perception encoder $F_{mono}$, the surround refinement model $F_{fusion}$, and the detector $D$ to accurately predict $Y^i$: $Y^i=D(F_{fusion}(F_{mono}(X^i),K^i,E^i))$.

\subsection{Surround Refinement Degradation}

\begin{figure}[t]
    \centering
    \includegraphics[width=1\linewidth]{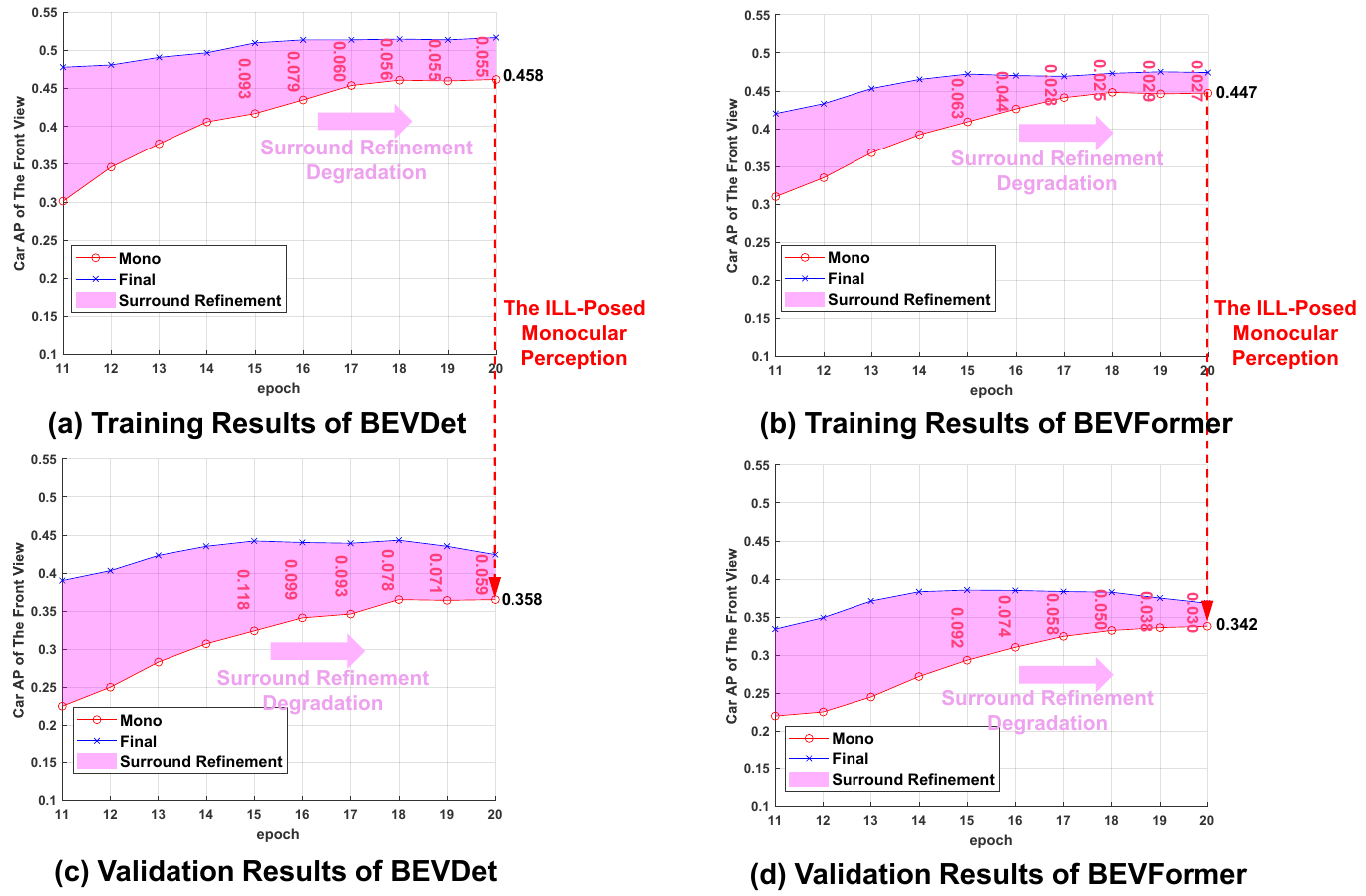}
    \caption{Visualization of surround refinement capability. The pink zone (surround refinement ability of $F_{fusion}$) is the disparity between monocular results (red line) and final fusion results (blue line). BEVDet and BEVFormer were jointly trained on nuScenes, Waymo, and Lyft datasets. The Average Precision (AP) results of monocular and fusion outcomes are reported on the Lyft dataset. For brevity, we report only the 'Car' category results in the front view.}
    \vspace{-4mm}
    \label{Fig:ana}
\end{figure}

The previous approach used additional information to supervise the monocular perception model $F_{mono}$ and further used information from $F_{mono}$ to prompt the surround refinement model $F_{fusion}$~\cite{jiang2023far3d,wang2023lift2d,yang2023bevformerv2}. However, monocular depth perception is an ill-posed problem, and monocular supervision only makes $F_{mono}$ overfit camera parameters and scenes. To make matters worse, the surround refinement model $F_{fusion}$ relies heavily on monocular accurate depth information to fuse multi-view features during training, which essentially prevents $F_{fusion}$ from learning the surround refinement abilities. To elucidate, BEVDet and BEVFormer were jointly trained on multi-datasets. 
nuScenes, Waymo, and Lyft datasets as depicted in Fig.~\ref{Fig:ana}. 



Fig.~\ref{Fig:ana} illustrates four objective phenomena for both BEVDet and BEVFormer: (1) During training, the red line (the performance of monocular perception) gradually approaches the blue line (the performance of the final result). (2) During training, the surround refinement stage (pink area) gradually reduces with a smaller contribution to the final result. (3) Compared with the training and validation results, the monocular (red line) results decreased significantly, for example, the monocular AP result of BEVDet is 0.487 in the training set and 0.374 in the validation set under the 20th epoch. (4) Compared with the training and validation results, the surround refinement (pink area) increased relatively or stayed the same, for example, the increased AP result (pink area) of BEVDet is 0.026 in the training set and 0.054 in the validation set under the 19th epoch.

According to phenomena (1) and (2), during training, the final (blue line) result gradually depends on the monocular perception ability (red line), and the ability of the surround refinement ability (pink area) is gradually weakened. According to phenomenon (3), the monocular perception performance will be dramatically reduced when facing different scenes or camera parameters because it's a naturally ill-posed problem. Phenomenon (4) shows that the surround refinement model has stronger generalization because it can use the multi-view information to refine the ill-posed monocular geometric representation. However, combined with phenomena (1), (2), (3), and (4), the stronger generalization of the surround refinement model is affected by overfitted monocular perception during training, called surround refinement degradation. Therefore, this paper aims to improve surround refinement ability while retaining strong monocular perception ability.

\section{Method}

\begin{figure}[t]
    \centering
    \includegraphics[width=1\linewidth]{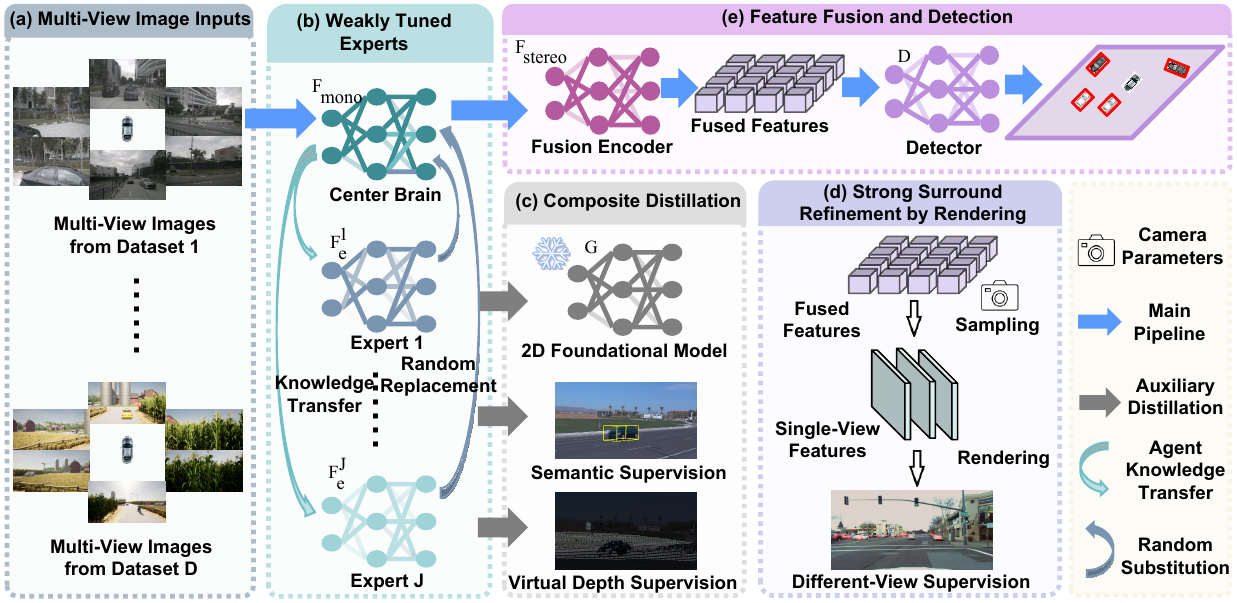}
    \caption{Our joint training framework for the MC3D-Det task. The blue arrows represent the general process of MC3D-Det: different datasets are fed into the monocular perception encoder (central brain $F_{mono}$) to extract preliminary features and then sent to the surround refinement model $F_{fusion}$; The final result is output through the detector. (b) weakly tuned experts distill part of the knowledge from the central brain. These experts replace parts of the sample in the central brain, which is a way of building noise features to augment the fusion module. (c) The composite knowledge of the generalized 2D foundational model, more fine-grained attributes, and standardized virtual depth are comprehensively distilled into the central brain. (d) Strong surround refinement by rendering is used to make the fusion model $F_{fusion}$ learn more robust geometric features.}
    \label{fig:framework}
    \vspace{-5mm}
\end{figure}

To enhance surround refinement ability, we design a plug-and-play MC3D-Det framework as shown in Fig.~\ref{fig:framework}. Specifically, the main pipeline (blue arrow) includes a monocular perception module ($F_{mono}$), fusion modules ($F_{fusion}$), and detection head ($D$), which is the typical representation of most MC3D-Det algorithms. Weakly tuned experts ($E_j$) are trained in different subsets, and the center brain ($F_{mono}$) is trained in all datasets. These weakly tuned experts are used to build ill-posed monocular features to reinforce the surround refinement ability of $F_{fusion}$. Then, the surround refinement rendering is used to constrain the fusion model ($F_{fusion}$) to learn more accurate geometric information. In addition, we use composite distillation to assist monocular perception in learning generalized and fine-grained features.

\subsection{Weak-to-Strong Eliciting}

The core of the surround refinement degradation is that the fusion model $F_{fusion}$ heavily depends on the overfitting depth information of $F_{mono}$ during training. An obvious solution is to construct some inaccurate monocular depth as a way of augmenting, which can assist the fusion network $F_{fusion}$ to use the multi-view features to refine geometric depth. To simulate depth bias facing new camera Settings and new scenes, we train weakly tuned experts on the divided training sets. Each expert has certain shortcomings with limited training sets, which can be used to produce ill-posed monocular features. Thus, the surround refinement model $F_{fusion}$ can learn to refine and fuse these ill-posed features from these experts.

\begin{figure}[t]
    \centering
    \begin{subfigure}{0.68\linewidth}
        \centering
        \includegraphics[width=\linewidth]{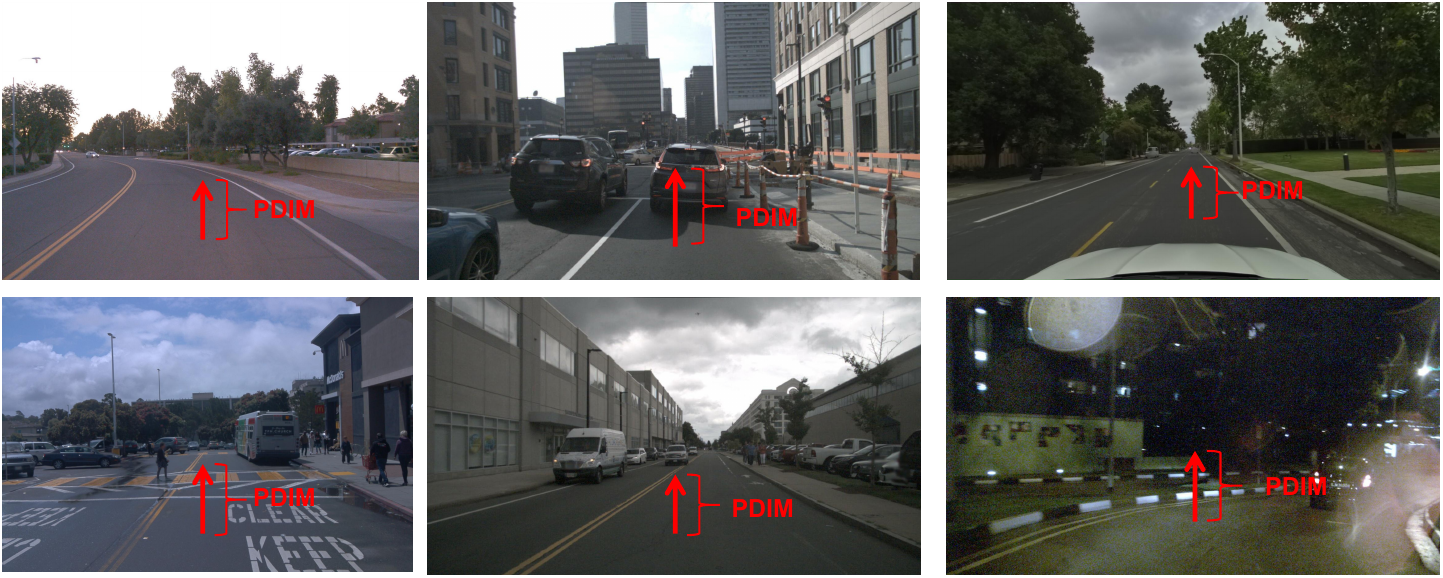}
        \caption{PDIR value visualization in different scenarios.}
        \label{Fig:PDIR0}
    \end{subfigure}
    \hfill
    \begin{subfigure}{0.30\linewidth}
        \centering
        \includegraphics[width=\linewidth]{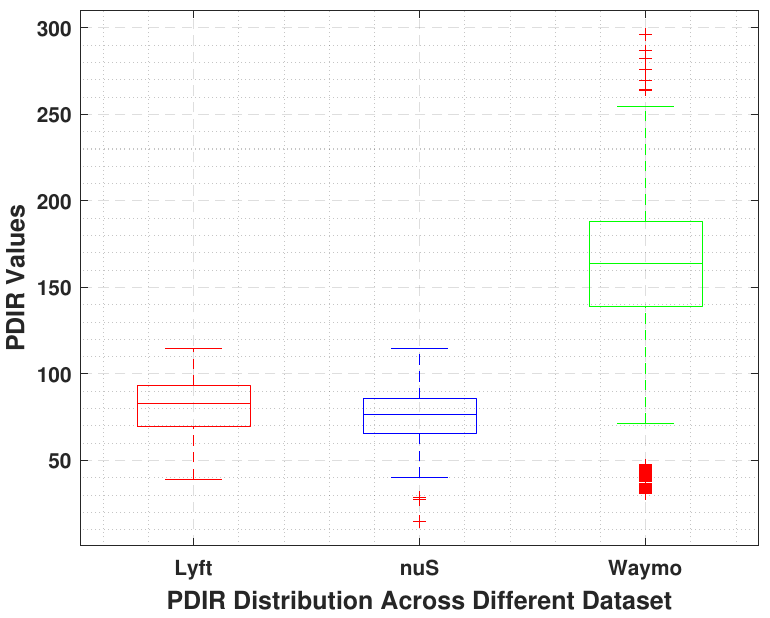}
        \caption{PDIR Distribution of different datasets.}
        \label{Fig:PDIR1}
    \end{subfigure}
    \caption{PDIR visualization. (a) PDIR reflects how far the ground depth changes over the pixels in the image. (b) PDIR has different distributions in the same dataset, which reflects not only the difference between different data sets but also the difference between specific scenes.}
    \vspace{-4mm}
    \label{Fig:PDIR}
\end{figure}

\textbf{Divided Training Set.} To build weakly tuned experts, the intuitive way is to build different training sets to train different experts. Multi-dataset joint training requires that MC3D-Det can better handle different camera intrinsic~\cite{Wang_2023_CVPR}, camera extrinsic (viewpoint)~\cite{klinghoffer2023Viewpoint,yang2023bevheight} and scene structure~\cite{li2023bev,xie2023benchmarking}. Based on this, we designed the Pavement Depth Increasing Rate $\text{PDIR}$ as shown in Fig.~\ref{Fig:PDIR}, which described the depth change rate of the image center line. Specifically, we project the object's box of $j$-th sample from 3D coordinates $P$ to the 2.5D front view $\hat{P}_{j}$ using the intrinsic $K'_j$ and extrinsic parameters $E'_j$: $\hat{P}_{j} = (ud, vd, d) = K'_j E'_j P$, $d$ represents the depth between the object and the view's optical center. Then the least square method calculates the plane representation $A\hat{ud}+B\hat{vd}+C\hat{d}+D=0$ of the ground surface based on the ground of the objects $\hat{P}_{j}^{ground}$. Then, we will project this plane to the front car view and calculate the number of pixels $\text{PDIR}$ corresponding to the center ($\hat{u}=0$) of the front view image $v_{max}(\hat{d_{min}})-v_{min}(\hat{d_{min}})$ meters, $v=\frac{-D-C\hat{d}}{B\hat{d}}$.

In theory, $\text{PDIR}$ is linearly proportional to the focal length of the camera because the depth information $\hat{d}$ is related to the camera focal. The placement of the camera (height or angle) and the pavement (height variation) will also affect parameter $\text{PDIR}$ because the plane equation $B, C, D$ is related to the camera pavement and the pavement.

\textbf{Weakly Tuned Experts.} Based on different training subsets, weakly tuned experts can be trained. Specifically, we set up a central brain $F_{mono}$ and multiple weakly tuned experts $F_{e}^{j}$. The central brain $F_{mono}$ is trained on all data. The weakly tuned  experts $F_{e}^{j}$ are trained by only distilling the knowledge from the central brain $F_{mono}$:
\begin{equation}
\begin{split}
    \mathcal{L}_{ag}^{j}=\sum_i^N  W^j_i (1-cos(F_{e}^{j}(I_i),F_{mono}(I_i))),
    \label{FOV_expert_0}
\end{split}
\end{equation}
where $\mathcal{L}_{ag}^{j}$ represents the distillation loss of $j$-th expert. By weighting different samples according to $\text{PDIR}$, $\mathcal{L}_{ag}^{j}$ makes the knowledge of different experts uneven. To simplify the problem, we only use 2 experts in this paper to prove the effectiveness of the method. So, the degree of distillation of different experts is $W^1_i=\frac{\exp(PDIR/PDIR_{Max})}{\sum_i^N \exp(PDIR/PDIR_{Max})}$ and $W^2_i=\frac{exp((PDIR_{Max}-PDIR)/PDIR_{Max})}{\sum_i^N \exp((PDIR_{Max}-PDIR)/PDIR_{Max})}$. The dynamic weighting of different samples allows different experts to have different abilities for different sub-datasets.

\textbf{Strong Surround Refinement by Rendering.} Each expert has certain shortcomings with limited training sets, which can be used to produce ill-posed monocular features. We used a straightforward strategy, which is to randomly replace the image features of central brain $F_{mono}$ with the ill-posed features of an expert with the probability of $p$. Then, the surround refinement model $F_{fusion}$ can learn to refine geometric features against the ill-posed features of weakly tuned experts. To further improve the surround refinement ability, we use a rendering strategy to constrain instance objects that are modified to fit in a reasonable BEV space. Following~\cite{lu2023pdbev}, implicit rendering, a plug-and-play constraint $\mathcal{L}_{render}$, is used to enhance surround refinement.

\subsection{Composite Distillation} 
With the weak-to-strong eliciting mechanism, we can enhance monocular perception without surround refinement degradation. The foundation model has achieved remarkable results because it learns more robust space features with large-scale 2D data~\cite{kirillov2023sam,zhao2023fastsam,caron2021dino,oquab2023dinov2,radford2021clip}. Intuitively, distilling robust 2D features assists in effectively improving capabilities for other tasks~\cite{wang2023lift2d,jiang2023far3d,yang2023bevformerv2}. 


\textbf{2D Semantic distillation.} MC3D-Det needs not only generalized semantic features but also geometric features and fine-grained attribute features. So the semantic features of the 2D foundational model are only distilled into the part of the channel of the image extractor $F_{mono}$. At the same time, different 2D pre-trained models $G_{pre}$ differ from our image extractors $F_{mono}$ in feature resolution and channel number, so we use a two-layer convolution head $P$ to project to the same size:
\begin{equation}
\begin{split}
    \mathcal{L}_{dis}= \sum_i^N (1-cos(P(G_{pre}(I_i)), M(F_{mono}(I_i)))),
    \label{FOV_SG_0}
\end{split}
\end{equation}
where mask $M$ is to use the first $K$ channels of $F_{mono}$ for distillation, $N$ is the image numbers. It is worth mentioning that $G_{pre}$ will not be updated, $P$ will be updated. Besides, these 2D foundational models cannot provide some fine-grained and task-specific attributes (depth, length, width, and height). Then, we build an auxiliary network to infuse this additional knowledge into $F_{mono}$. Following~\cite{centerpoint}, focal loss $Focal()$~\cite{focal} and L1 loss $L1$ are used to supervise the class information and object dimensions of the auxiliary network.


\textbf{Virtual camera supervision.} Depth information is integrated into monocular understanding encoder $F_{mono}$ in advance, which quickly helps the algorithm initialize the object position in the BEV plane~\cite{li2022bevdepth,FBBEV}. However, camera parameters are different across different datasets, and depth is affected by focal length~\cite{yin2023metric3d,Wang_2023_CVPR,yin2023metric3d,lu2023pdbev}. Following~\cite{Wang_2023_CVPR}, we use virtual camera depth $\mathcal{L}_{pg}$ here to establish monitoring of monocular depth.


\subsection{Dataset Merge Strategy} 
\label{sec:tr}
Most MC3D-Det are single-dataset training-and-testing paradigms, and training and testing are the same camera parameters and the camera number. However, the camera parameters and number are different across datasets. To this end, we recalculate the actual intrinsic camera parameters input to the network according to the pinhole imaging principle~\cite{yin2023metric3d}. To solve the problem of different numbers of cameras, we introduced a ghost camera for the missing camera data set, for example, Waymo only has five cameras while other datasets have six cameras. This ghost camera has a focal length of 0, and the pixel values in its images are all set to 0. This approach can be effectively applied to nearly all MC3D-Det algorithms and has demonstrated no adverse effects in our experimental evaluation.



\section{Experiment}

\subsection{Datasets} 

We evaluate our proposed approach on three datasets, including one virtual and two real autonomous driving datasets: nuScenes~\cite{caesar2020nuscenes}, Waymo~\cite{sun2020waymo}, Lyft~\cite{lyft} and DeepAccident~\cite{wang2023deepaccident}. Different datasets have different camera parameters and environments~\cite{zheng2023cross}. In addition, different datasets have adopted inconsistent category and granularity definitions. We classify all different vehicles as "vehicles," motorcycles and bicycles as "two-wheelers," and different people as "pedestrians."

\subsection{Evaluation Metrics}  Each category reports AP and NDS$^+$metrics. AP is officially exactly the same as nuScenes, that is, calculated in the BEV perspective. NDS$^+$ is an extension of the official NDS metric used in nuScenes. Since the velocity labels are not directly comparable in different datasets, NDS$^+$ contains the Average Precision (AP), the Average Translation Error (ATE), the Average Scale Error (ASE), and the Average Orientation Error (AOE):
\begin{equation}
\setlength{\abovedisplayskip}{2pt}
\setlength{\belowdisplayskip}{3pt}
  \rm{NDS^+ = \frac{1}{6}[3 \, AP + \sum_{TP \in ATE, ASE, AOE}(1-min(1,TP))]},
\end{equation}
Then we calculate the mean of AP and NDS$^+$ for the three categories to get mAP and mNDS$^+$. It is worth noting that all metrics are reported on the validation in the range [-50m, 50m] along the x-y axes of the BEV plane.

\subsection{Implementation Details}
We train our models using the AdamW optimizer, with a gradient clip and a learning rate of 2e-4. We use a total batch size of 24 on 8 Tesla 3090 GPUs. Training rounds are 20. The final image input resolution for all datasets is set to $704 \times 384$. All image feature encoders are ResNet50 pre-trained in ImageNet. We apply data augmentation techniques such as random flipping, random scaling, and random rotation within a range of $[-5.4^\circ, 5.4^\circ]$ to the original images during training. It is worth noting that the random scaling is in a range of $[-0.06, 0.11]$, which affects the actual parameters of the cameras. Dataset merge strategy are used on all the baselines as discussed at Sec.~\ref{sec:tr}. All algorithms use these parameters by default to ensure fairness. All algorithms are trained on the training set and tested on the validation set.

\subsection{Multiple dataset joint training benchmark}

\begin{table*}[t]
\vspace{-0.1cm}
  \centering
  \begin{tabular}{c|c|cccc}
    \toprule
    Method & Test dataset & car & two-wheeler & pedestrian & mAP/mNDS$^+$\\
    \midrule
     Oracle&nuScenes&0.385/0.508&0.175/0.249&0.205/0.247&0.255/0.335\\
     DETR3D~\cite{wang2022detr3d}&nuScenes&0.111/0.282&0.016/0.121&0.067/0.149&0.065/0.184\\
     PETR~\cite{liu2022petr}&nuScenes&0.339/0.456&0.129/0.180&\underline{0.231}/0.244&0.233/0.294\\
     BEVDet~\cite{huang2021bevdet}&nuScenes&0.392/0.515&0.169/0.254&0.193/0.238&0.251/0.336\\
     BEVFormer~\cite{li2022bevformer}&nuScenes&0.325/0.294&0.115/0.146&0.170/0.174&0.203/0.205\\
     BEVDepth~\cite{li2022bevdepth}&nuScenes&\underline{0.411}/\underline{0.537}&\underline{0.176}/\underline{0.255}&0.213/\underline{0.258}&\underline{0.267}/\underline{0.350}\\
     FB-BEV~\cite{FBBEV}&nuScenes&0.340/0.366&0.127/0.168&0.186/0.181&0.218/0.238\\
     Ours&nuScenes&\textbf{0.447}/\textbf{0.540}&\textbf{0.185}/\textbf{0.258}&\textbf{0.246}/\textbf{0.275}&\textbf{0.293}/\textbf{0.358}\\
     \midrule
     Oracle&Waymo&0.426/0.536&0.135/\textbf{0.265}&0.334/0.328&0.298/\underline{0.376}\\
     DETR3D~\cite{wang2022detr3d}&Waymo&0.173/0.325&0.033/0.134&0.142/0.191&0.116/0.217\\
     PETR~\cite{liu2022petr}&Waymo&0.360/0.469&0.142/0.211&0.262/0.265&0.255/0.315\\
     BEVDet~\cite{huang2021bevdet}&Waymo&\underline{0.437}/\underline{0.543}&\underline{0.150}/0.249&0.339/0.323&\underline{0.307}/0.372\\
     BEVFormer~\cite{li2022bevformer}&Waymo&0.353/0.332&0.100/0.143&0.204/0.204&0.219/0.226\\
     BEVDepth~\cite{li2022bevdepth}&Waymo&0.432/0.536&0.144/0.250&\underline{0.342}/\underline{0.332}&0.306/0.373\\
     FB-BEV~\cite{FBBEV}&Waymo&0.363/0.383&0.122/0.186&0.220/0.201&0.235/0.257\\
     Ours&Waymo&\textbf{0.466}/\textbf{0.549}&\textbf{0.157}/\underline{0.261}&\textbf{0.355}/\textbf{0.341}&\textbf{0.326}/\textbf{0.384}\\
     \midrule
     Oracle&Lyft&0.408/0.530&0.349/0.440&0.114/0.199&0.290/0.390\\
     DETR3D~\cite{wang2022detr3d}&Lyft&0.078/0.306&0.055/0.170&0.005/0.109&0.046/0.195\\
     PETR~\cite{liu2022petr}&Lyft&0.323/0.455&0.227/0.332&0.105/0.166&0.218/0.318\\
     BEVDet~\cite{huang2021bevdet}&Lyft&0.457/0.580&\underline{0.353}/\underline{0.441}&0.109/0.199&0.306/0.407\\
     BEVFormer~\cite{li2022bevformer}&Lyft&0.385/0.410&0.289/0.253&0.092/0.153&0.255/0.272\\
     BEVDepth~\cite{li2022bevdepth}&Lyft&\underline{0.473}/\underline{0.581}&0.338/0.436&\underline{0.126}/\underline{0.201}&\underline{0.312}/\underline{0.406}\\
     FB-BEV~\cite{FBBEV}&Lyft&0.406/0.455&0.268/0.331&0.104/0.159&0.259/0.315\\
     Ours&Lyft&\textbf{0.505}/\textbf{0.603}&\textbf{0.355}/\textbf{0.453}&\textbf{0.140}/\textbf{0.212}&\textbf{0.333}/\textbf{0.422}\\
    \bottomrule
  \end{tabular}
  \caption{Comparison of different approaches on validation set under multi-dataset joint training. The \textbf{best} and \underline{second best} scores of each metric are highlighted in \textbf{bold} and \underline{underline}, respectively.}
  \vspace{-4mm}
  \label{tab:all-result}
\end{table*}

In combining three real data sets, we evaluated DETR3D~\cite{wang2022detr3d}, PETR~\cite{liu2022petr}, BEVFormer~\cite{li2022bevformer}, BEVDet~\cite{huang2021bevdet}, BEVDepth~\cite{li2022bevdepth} and FB-BEV~\cite{FBBEV} as shown in Tab.~\ref{tab:all-result}. All algorithms are trained on nuScenes, waymo, and lyft training sets, and tested on validation sets. All methods use the dataset merge strategy described in Sect.~\ref{sec:tr}, otherwise, it would not be possible to train together due to the different number of cameras and defects in the native code. Oracle is the result of separate training and evaluation on each dataset using BEVDet. Our method is plug-and-play, since it is mainly applied to image feature extractors. Considering that BEVDet is simple and efficient, our proposed framework is instantiated from BEVDet. 

As shown in Tab.~\ref{tab:all-result}, DETR3D performs poorly because it doesn't take camera parameters as input to the network at all. PETR takes 3D position-coded information as input, so it can present acceptable results on multiple datasets. BEVFormer uses the camera parameter to query potential image features in the form of acceptable results. BEVDet and BEVDepth achieve satisfactory results because they explicitly decoupage camera parameters rather than rely on learning. Combining multiple data sets makes BEVDet surpass Oracle, which shows that the combined training of multiple data sets can effectively improve the model capability. Our method is applied to BEVDet to enhance the surround refinement ability, which makes the model more generalized to different camera parameters and scenes.

\subsection{Sim-Real joint training}

The virtual engine has better controllability and can generate various scenarios and samples: different weather and scenes~\cite{shift2022}, corner case~\cite{Crash,Wang_2023_DeepAccident}, vehicle-to-everything~\cite{OPV2V,V2X-Sim}. Therefore, breaking the domain gap between virtual and real datasets can further facilitate the closed-loop form of visually-oriented planning~\cite{jia2023driveadapter,sima2023drivelm,wang2023drivemlm,yuan2023ad}. To evaluate the perception ability of the model in both virtual and real scenarios, we establish a benchmark for the sim-real joint training as shown in Tab.~\ref{tab:real-sim}.

\begin{table*}[t]
\vspace{-0.1cm}
  \centering
  \begin{tabular}{c|c|cccc}
    \toprule
    Method& Test dataset & car & two-wheeler & pedestrian &  mAP/mNDS$^+$ \\
    \midrule
     Oracle&nuScenes&\underline{0.385}/\textbf{0.508}&\underline{0.175}/\underline{0.249}&\underline{0.205}/\underline{0.247}&\underline{0.255}/\underline{0.335}\\
     DETR3D~\cite{wang2022detr3d}&nuScenes&0.119/0.262&0.023/0.120&0.059/0.145&0.067/0.176\\
     PETR~\cite{liu2022petr}&nuScenes&0.311/0.298&0.143/0.254&0.235/0.262&0.297/0.271\\
     BEVDet~\cite{huang2021bevdet}&nuScenes&0.368/0.470&0.156/0.241&0.235/0.267&0.253/0.326\\
     BEVFormer~\cite{li2022bevformer}&nuScenes&0.304/0.290&0.136/0.169&0.179/0.183&0.206/0.214\\
     BEVDepth~\cite{li2022bevdepth}&nuScenes&0.329/0.453&0.139/0.238&0.223/0.251&0.230/0.314\\
     Ours&nuScenes&\textbf{0.394}/\underline{0.487}&\textbf{0.184}/\textbf{0.253}&\textbf{0.247}/\textbf{0.273}&\textbf{0.275}/\textbf{0.338}\\
     \midrule
     Oracle&DeepAccident&\textbf{0.319}/\textbf{0.399}&\textbf{0.165}/\textbf{0.303}&\underline{0.191}/\underline{0.252}&\textbf{0.225}/\textbf{0.308}\\
     DETR3D~\cite{wang2022detr3d}&DeepAccident&0.041/0.169&0.003/0.165&0.016/0.111&0.020/0.148\\
     PETR~\cite{liu2022petr}&DeepAccident&0.239/0.345&0.116/0.182&0.151/0.188&0.168/0.238\\
     BEVDet~\cite{huang2021bevdet}&DeepAccident&0.281/0.368&0.143/0.257&0.185/0.243&0.203/0.289\\
     BEVFormer~\cite{li2022bevformer}&DeepAccident&0.225/0.240&0.112/0.162&0.134/0.160&0.157/0.188\\
     BEVDepth~\cite{li2022bevdepth}&DeepAccident&0.247/0.328&0.131/0.244&0.172/0.230&0.183/0.267\\
     Ours&DeepAccident&\underline{0.316}/\underline{0.374}&\underline{0.155}/\underline{0.290}&\textbf{0.195}/\textbf{0.258}&\underline{0.222}/\underline{0.307}\\
    \bottomrule
  \end{tabular}
  \caption{Comparison of different approaches on Real-Sim joint training (nuScenes and DeepAccident). The \textbf{best} and \underline{second best} scores of each metric are highlighted in \textbf{bold} and \underline{underline}, respectively.}
  \label{tab:real-sim}
  \vspace{-4mm}
\end{table*}

As shown in the Tab.~\ref{tab:real-sim}, our algorithm is better than all the existing algorithms. On nuScenes validation set, it is worth mentioning that BEVDet is improved in the combined virtual-real training process compared to Oracle in the two categories of two-wheeler and pedestrian, which indicates that virtual data can be helpful for real scenarios in difficult and fewer targets. BEVDepth is much worse than BEVDet, which indicates that there is a clear domain gap between the depth of information provided by virtual and real point clouds. On the virtual dataset, our algorithm has almost the same performance as Oracle, indicating that our algorithm can effectively mitigate the gap between virtual and real. Combined virtual-real training increases the performance of real data sets, but decreases the performance of virtual data sets. Because real-world scenarios can be more complex or diverse, models are forced to learn more generalized feature representations. At the same time, the virtual scene will provide more scenes that can improve the performance of the real scene.  

\subsection{Ablation study}

To further verify the effectiveness of our algorithm, we conducted ablation study on our algorithm. We evaluated each of the three sections: unevenly-skelled expert (USE), surround refinement enhancement (SRE), and composite distillation (CD). We trained on three real datasets (nuScence, Waymo, Lyft) and tested on validation sets as shown in Tab.~\ref{tab:ablation}. Each module can effectively improve the performance on all validation sets. When combining the three modules, significant increases on each metric can be achieved. In addition, we can observe that although the CD method incorporating monocular supervision can lead to surround refinement capability loss, the performance is still improved compared to without this method, which indicates that our USE can effectively improve the surround refinement ability. 

\begin{table}[t]
  \centering
  \scalebox{0.85}{
  \begin{tabular}{ccc|cc|cc|cc}
    \toprule
    \multicolumn{3}{c|}{} &\multicolumn{2}{c|}{nuScence}& \multicolumn{2}{c|}{Waymo}& \multicolumn{2}{c}{Lyft}\\
    \midrule
    USE & SRE & CD & mAP $\uparrow$ & NDS$^+$ $\uparrow$ & mAP $\uparrow$ & NDS$^+$ $\uparrow$ & mAP $\uparrow$ & NDS$^+$ $\uparrow$ \\
    \midrule
    &    &                              &0.251 & 0.336  &0.307 & 0.372&0.306 & 0.407\\
   \checkmark &  &                      &0.269 & 0.342  &0.317 & 0.378&0.318 & 0.414 \\
    & \checkmark &                      &0.258 & 0.349  &0.311 & 0.375&0.310 & 0.411 \\
    &  & \checkmark                     &0.264 & 0.343  &0.314 & 0.376 &0.315 & 0.412\\
   \midrule
   \checkmark & \checkmark & \checkmark &\textbf{0.293} & \textbf{0.358}  &\textbf{0.326} & \textbf{0.384} &\textbf{0.333} & \textbf{0.422}  \\
    \bottomrule
  \end{tabular}
  }
  \caption{Ablation study of different modules of Scale-BEV. USE means using an unevenly-skelled expert strategy. SRE stands for surround refinement enhancement strategy. CD stands for composite distillation. The \textbf{best} scores of each metric are highlighted in \textbf{bold}.}
  \label{tab:ablation}
  \vspace{-6mm}
\end{table}

\begin{table}[t]
    \centering
    \begin{subtable}{0.45\textwidth}
        \centering
        \scalebox{0.85}{
            \begin{tabular}{c|cc}
                \toprule
                nuScenes & mAP $\uparrow$ & mNDS$^+$ $\uparrow$ \\
                \midrule
                Baseline     &0.251 & 0.336 \\
                DS      &0.256 & 0.337 \\
                RD      &0.254 & 0.238 \\
                PDIR    &0.269 & 0.342 \\
                \bottomrule
            \end{tabular}
        }
        \caption{Comparison of training set splitting strategies. }
        \label{tab:tss}
    \end{subtable}
    \hfill
    \begin{subtable}{0.45\textwidth}
        \centering
        \scalebox{0.90}{
            \begin{tabular}{c|cc}
                \toprule
                nuScenes & mAP $\uparrow$ & mNDS$^+$ $\uparrow$ \\
                \midrule
                CLIP     &0.286 & 0.351 \\
                SAM      &0.295 & 0.359 \\
                Fast-SAM &0.293 & 0.358 \\
                DINO     &0.296 & 0.356 \\
                DINO-v2  &0.298 & 0.358 \\
                \bottomrule
            \end{tabular}
        }
        \caption{Comparison using different 2D foundation models. }
        \label{tab:2d}
    \end{subtable}
    \vspace{-4mm}
    \caption{Discussion of dataset partitioning strategy and 2D foundation model. }
    \vspace{-4mm}
\end{table}

\subsection{Further analysis}

To further discuss the details of the algorithm and in-depth understanding. We had a discussion:

\noindent \textbf{Training set splitting.} The core of building different weakly tuned experts is to divide different training subsets. One or two straightforward solutions are (1) a dataset as a subset (DS); (2) Random division (RD). As is evident in Tab.~\ref{tab:tss}, our proposed strategy PDIR has made a significant improvement, which shows that our partitioning method can reflect both the camera parameters of the camera as well as the scene structure. In addition, the other two methods are also effective and this illustrates the effectiveness of our weakly tuned expert strategy.


\noindent \textbf{2D foundmental model.} We tested the effect of different 2D foundmental models for distillation, including CLIP~\cite{radford2021clip}, SAM~\cite{kirillov2023sam}, Fast-SAM~\cite{zhao2023fastsam}, DINO~\cite{caron2021dino}, and DINO-v2~\cite{oquab2023dinov2}. As shown in Tab.~\ref{tab:2d}, all base models performed similarly, except the CLIP. This is because the fine-grained features of CLIP are poor compared to the others.


\noindent \textbf{Dataset merge strategy.} The number of cameras varies from dataset to dataset, and we use a strategy to make the focal length infinitely small. To prove that this strategy is harmless, we trained PETR, BEVDet, and BEVFormer on Waymo in two ways, one using only five cameras, and the other using the Ghost camera to complete six cameras. The Tab.~\ref{tab:tr} shows that our strategy has almost no effect on the final results. This strategy is necessary for joint training of different camera numbers, which is used for all MC3D-Det algorithms.

\begin{table}[t]
    \centering
    \vspace{-2mm}
    \begin{subtable}{0.40\textwidth}
        \centering
        \scalebox{0.75}{
            \begin{tabular}{c|cc|cc}
                \toprule
                \multicolumn{1}{c|}{} &\multicolumn{2}{c|}{Original}& \multicolumn{2}{c}{Ghost Camera}\\
                \midrule
                nuScenes & mAP $\uparrow$ & mNDS$^+$ $\uparrow$ & mAP$\uparrow$ & mNDS$^+$ $\uparrow$\\
                \midrule
                PETR      &0.249 & 0.309  &0.245 & 0.307 \\
                BEVDet    &0.298 & 0.376  &0.301 & 0.377 \\
                BEVFormer &0.224 & 0.246  &0.243 & 0.242 \\
                \bottomrule
            \end{tabular}
        }
        \caption{The effects of ghost camera.}
        \label{tab:tr}
    \end{subtable}
    \hfill
    \begin{subtable}{0.50\textwidth}
        \centering
        \scalebox{0.80}{
            \begin{tabular}{c|cc|cc}
                \toprule
                \multicolumn{1}{c|}{} &\multicolumn{2}{c|}{w/o ours}& \multicolumn{2}{c}{w ours}\\
                \midrule
                nuScenes & mAP $\uparrow$ & mNDS$^+$ $\uparrow$ & mAP$\uparrow$ & mNDS$^+$ $\uparrow$\\
                \midrule
                PETR      &0.251 & 0.336  &0.293 & 0.358 \\
                BEVDet    &0.251 & 0.336  &0.293 & 0.358 \\
                BEVFormer &0.203 & 0.205  &0.257 & 0.341 \\
                \bottomrule
            \end{tabular}
        }
        \caption{The plug-and-play capability testing of our method.}
        \label{tab:pp}
    \end{subtable}
    \vspace{-4mm}
    \caption{Discussion of ghosting cameras and plug and play capabilities.}
    \vspace{-8mm}
\end{table}


\noindent \textbf{Plug-and-play ability.} Our approach is plug-and-play and can be applied to most MC3D-Det algorithms, and we apply it here to the most common three: PETR, BEVDet, and BEVFormer. As shown in Tab.~\ref{tab:pp}, our algorithm can improve performance effectively when applied to different MC3D-Det paradigms, which also indicates the strong plug-and-play capabilities of our algorithm.


\section{Summary}
In this paper, we aimed to expand the training scale for the MC3D-Det task, and at the same time utilize the 2D foundation model. For multi-dataset joint training, the most important thing is to deal with different camera parameters and environments. However, we find that the surround refinement ability decreases gradually because the multi-view fusion model depends on the monocular depth perception during training. When it came to the test phase, the monocular could not accurately handle different camera parameters and scenes, and the surround refinement ability was not fully developed in the training phase. For this purpose, we propose a plug-and-play training framework for MC3D-Det. Our method can effectively improve the performance of multi-dataset joint training. We have established a large-scale benchmark. In particular, our paper also explores the joint training of virtual and reality, to explore the possibility of closed-loop end-to-end training. Our algorithm is plug-and-play and has shown effectiveness on multiple MC3D-Det algorithms.

%
%
\bibliographystyle{splncs04}
\bibliography{egbib}
\end{document}